\documentclass[journal]{IEEEtran}

\ifCLASSINFOpdf
  \usepackage[pdftex]{graphicx}
\else
  \usepackage[dvips]{graphicx}
\fi

\usepackage{amsmath}
\usepackage{algorithmic}
\usepackage{array}
\usepackage{url}

\usepackage{caption}
\usepackage{subfig}
\usepackage{comment}

\usepackage{csquotes}

\usepackage{todonotes}
\usepackage{color}

\title{
Who Killed Albert Einstein? \\
From Open Data to Murder Mystery Games
}
\author{Gabriella A. B. Barros$^{1}$, Michael Cerny Green$^{1}$, Antonios Liapis$^{2}$ and Julian Togelius$^{1}$ \\
1: Tandon School of Engineering,
New York University,
New York, USA\\
2: Institute of Digital Games,
University of Malta,
Msida, Malta\\
gabbbarros@gmail.com, mcgreentn@gmail.com, antonios.liapis@um.edu.mt, julian@togelius.com
}

\begin{document}

\maketitle

\begin{abstract}
This paper presents a framework for generating adventure games from open data. Focusing on the murder mystery type of adventure games, the generator is able to transform open data from Wikipedia articles, OpenStreetMap and images from Wikimedia Commons into WikiMysteries. Every WikiMystery game revolves around the murder of a person with a Wikipedia article, and populates the game with suspects who must be arrested by the player if guilty of the murder or absolved if innocent. Starting from only one person as the victim, an extensive generative pipeline finds suspects, their alibis, and paths connecting them from open data, transforms open data into cities, buildings, non-player characters, locks and keys and dialog options. The paper describes in detail each generative step, provides a specific playthrough of one WikiMystery where Albert Einstein is murdered, and evaluates the outcomes of games generated for the 100 most influential people of the 20th century.
\end{abstract}

\section{Introduction}\label{sec:introduction}

Games that cast the player in the role of a detective, where the gameplay and main challenge revolve around solving a crime or mystery, have been popular for many decades. Some games, such as \emph{Where in the World is Carmen Sandiego?} (Br{\o}derbund Software 1985), task the player with finding a fugitive criminal. Other games, such as \emph{Indiana Jones and the Fate of Atlantis} (LucasArts 1992) or the series of \emph{Tomb Raider} (Eidos 1996) and \emph{Uncharted} (Naughty Dog 2007), see the player embark on an adventure to solve ancient mysteries in the face of opposition from shadowy goons. It is common for these games to feature frequent in-game travel to exotic locales around the world to interact with colorful people and to gather clues, solve puzzles and overcome resistance.

These games often make heavy use of real-world locations, stories, items and characters to build their narrative. Authoring such games is complex, time-consuming and requires considerable skill. However, the fact that these games depend on so much real-world information, and that such information is freely available in structured or semi-structured form from resources such as Wikipedia and OpenStreetMap, suggests that it would be possible to somehow automatically incorporate real-world information in these games. Furthermore, murder mysteries and similar adventure games are often highly structured, suggesting the possibility of generating the game itself. But could we do this in practical manner, and if so, how? 
%
While research projects in AI-based game generation such as \emph{Game-o-Matic} \cite{treanor2012gameomatic} and \emph{A Rogue Dream} \cite{cook2014aroguedream} create simple arcade-like and rogue-like games respectively, a murder mystery requires a much larger volume and variety of content such as locations, people, dialog and clues for solving the mystery. Moreover, \emph{consistency} of the content is necessary, both internally within the game narrative and externally as fidelity with the real world. The contribution of a generator of murder mysteries from open data is three-fold: (a) it explores how disparate data can be connected together to create and represent plot-lines, (b) it identifies a design ``formula'' and structure for murder mysteries and their constituent elements (such as dialog) which can be used to generate  mystery games from any Wikipedia entry, (c) it  tests the limits of autonomous game generation and issues that arise from absurd or incomplete source data (or from their algorithmic combination).


This paper presents \emph{WikiMystery}, a framework for generating complete, playable point-and-click adventure games with minimal human input: in this case, the name of a person who has a Wikipedia page. The WikiMystery generative system featured in this paper builds upon earlier work \cite{barros2016murder}, extending it significantly with a description of the full generative pipeline, a more sophisticated dialog system and a broader evaluation of nearly 100 games for solving the murder of the most influential people of the 20th century. The WikiMystery game generator is built on the previous project \emph{Data Adventures} \cite{barros2015dataadventures} and reuses much of that technology to discover paths between victim and suspects of the murder. 
The current framework however offers a much more engaging, coherent and complete experience with a clear goal to arrest the culprit of a murder. This is facilitated by extensive story branching towards several suspects, enhanced ludic elements as game objects that unlock certain locations, and enriched dialog elements that allow Non-Player Characters (NPCs) to share facts both about the mystery and about themselves (based on open data).

The paper starts with a brief survey in Section \ref{sec:background} on game, plot and dialog generation. The paper provides an overview of the generative pipeline of WikiMystery in Section \ref{sec:overview} and the specifics of culprit and evidence selection in Section \ref{sec:suspects}, path generation in Section \ref{sec:paths}, location, NPC and item generation in Section \ref{sec:enriching} and finally NPC dialog generation in Section \ref{sec:dialogue}. To assess the generated games, a sample playthrough is described in Section \ref{sec:playthrough}, while Section \ref{sec:evaluation} analyzes the games created for murders of the 100 most influential people of the 20th century. The paper leads to a discussion in Section \ref{sec:discussion} 
and concludes with Section \ref{sec:conclusion}.

\section{Background}\label{sec:background}

The WikiMystery system is a framework for transforming open data into adventure games. This section discusses the domains of data games, game plot, and dialog generation.

\subsection{Data Games}\label{sec:background_datagames}
In a world of ever-more ubiquitous technology, the amount of data we consume daily is rapidly increasing. 
One data category that is growing exponentially is open data, i.e.~information that can be freely used, re-used, and redistributed by anyone. Creating games out of data can be seen as a form of visualization, where instead of using charts or figures to make the information easier to grasp, one creates playable media.


Data games use real world information (such as open data) to automatically generate game content \cite{friberger2013data}. Players interact with this content during gameplay, and more often than not must learn how to understand the data in order to play the game well. Typically, to use data as game content one must select what parts of the data are useful for content generation, and structurally transform it into applicable game content. Some example data games are discussed below.

Open Trumps \cite{cardona2014open} is a data game where the cards' content is based entirely on published governmental data. Its generator creates a balanced Top Trumps deck using evolutionary algorithms. While it is not required to learn the data, it helps when playing Open Trumps. In MuseumVILLE\footnote{MuseumVILLE: (\url{https://github.com/bogusjourney/museumville/})}, content is selected from Europeana\footnote{Europeana is a portal for accessing digitised cultural heritage material, such as paintings and books, from more than 2,000 institutions across Europe.}, and the user, playing as a museum curator, must theme their museum based of their interests. BarChartBall ~\cite{togelius2013bar} is a physics game that uses UK census data to transform the playable level. It is necessary to infer how the data would affect the playfield, which is modified based upon how high or low the selected attribute was. A Rogue Dream ~\cite{cook2014aroguedream} uses the auto-complete results of Google queries (using templates) to choose names for player abilities, enemies and healing items; these names are then given a visual found via Google image search. Finally, geographical data from OpenStreetMap have been used to generate maps and players' initial positions for FreeCiv~\footnote{FreCiv is an open souce version of Civilization.  (\url{http://www.freeciv.org/})}~\cite{barros2015balanced}.


\subsection{Story, Quest and Dialog Generation in Games}\label{sec:background_story}

Research on story generation tends to focus on the textual form. In BRUTUS~\cite{bringsjord2000artificial}, 
the generator creates dark stories about characters with backgrounds and narratives.
In MINSTREL~\cite{Turner:1993:MCM:166478}, the generator uses author-level problem solving to write short stories based on King Arthur and his knights. Bardiche~\cite{vink2015bardiche} acts as a collaborative tool to create \textit{good} stories based on user input, improvising based on user input.
Creating games from stories, or stories from games, is part of the broader subject of recontextualizing data from one medium into another. Examples include the transformation of levels in \emph{Sonancia} \cite{lopes2015sonancia} and text in \emph{Audio Metaphor} \cite{thorogood2012audio} into soundscapes, or news articles into games in \emph{ANGELINA} \cite{cook2014automating}.


In adventure games, stories and quests are crucial for progression and this needs to be taken into account when generating them. The game \emph{Charbitat} \cite{alderman2006charbitat} procedurally generates the environment as the player explores it, but it lacks a sense of progression. In order to anchor the player, a quest generator was introduced to Charbitat that uses key-lock mechanisms to advance in the game world \cite{ashmore2007quest}. Similarly, a two-tiered procedural generator was built for \emph{Mystery of Solaris}~\cite{lavenderadventures}, constructing maps and missions via two separate grammars. First, a mission graph is constructed containing all quest information. The mission graph is then taken as input to the map generator, which builds a map around the mission.
\emph{Symon} \cite{fernandez2012procedural}, a point-and-click adventure game, uses procedural content generation to create meaningful puzzles. The generation in \emph{Symon} was expanded into the \emph{Puzzle Dice System} \cite{fernandez2014creating} for generating puzzles in adventure games. 

Since murder mystery games largely rely on interaction with non-player characters, the quality of NPC dialog is an important factor to gameplay experience. Dialog generation is hardly a new topic of research, going back all the way to \emph{ELIZA} \cite{weizenbaum1966eliza}. Another example is the \emph{Text2dialog} system \cite{hernault2008generating} which transforms monological text into dialog, and has agents act out the dialog. It uses textual coherence relationships to map text to question-answer pairs, and is able to create fairly believable dialog. However, we are less interested in creating realistic sequences of chat (and responding accommodatingly to any player request), but rather in driving NPC interactions towards a specific direction, i.e. providing clues to the player.
Interactive storytelling has studied dialog generation and delivery extensively, e.g. combining implicit forms of character expression, overall narrative goals, and emotional relationships between characters to generate realistic dialog \cite{cavazza2005dialogue}. 

\section{Overview of the Game and the Generator}\label{sec:overview}

WikiMystery is a data-based procedurally generated point-and-click adventure game. It uses data from Wikipedia, OpenStreetMap\footnote{OpenStreetMap is a open source project that attempts at mapping the world (\url{https://www.openstreetmap.org/}).} and Wikimedia Commons to automatically create different game content, from plot progressing to images. It was strongly inspired by classic adventure games such as \emph{Where in the World is Carmen Sandiego?} (Br{\o}derbund Software 1985). Fern\'{a}ndez-Vara and Osterweil \cite{vara2010adventure} describe some of the main characteristics of an adventure game: gameplay driven by the story, puzzle-solving core mechanics, interaction with the game world through object manipulation, and an in-world player-controlled character motivated to explore and interact with her surroundings.

In the game, the player assumes the role of a detective trying to solve a murder case. The victim is the central point of the story, and suspects are based on people related ---somehow--- to her. We use Wikipedia to identify possible suspects, out of which five are selected. The game plot is tree-structured: the victim is the root, each suspect is a leaf, and the path between them is a representation of the hyperlinks between the victim's and each suspect's Wikipedia articles.

Initially, the only location available is the victim's house, where the player can talk to people related to the victim. The player also becomes aware of who the five suspects for the murder are. As they interact with people inside the house, new locations, objects and NPCs become available. As the player explores and interacts with the world, they collect information about suspects' characteristics (e.g.~year of death, occupation etc.). Every suspect except the culprit has a value for one characteristic (e.g. ``died in 1980'' for suspect 1 or ``birth place in Cleveland'' for suspect 2) which acts as an evidence of innocence (identified in the dialog with a note ``... couldn't have done it.''). The player receives no such information about the culprit; the culprit also does not share the same value in a characteristic which is evidence of innocence for other suspects (i.e. the culprit did \emph{not} die in 1980 and was \emph{not} born in Cleveland). The game ends when the player issues an arrest warrant, identifying the culprit and specifying the values acting as evidence of innocence of the other suspects. If the player correctly finds the culprit and provides the correct evidence for the other suspects, the game is won. If the player does not specify the right culprit, or if the evidence for the innocence of any other suspect is incorrect, then the game is lost.

\begin{figure}[t]
\includegraphics[width=\columnwidth]{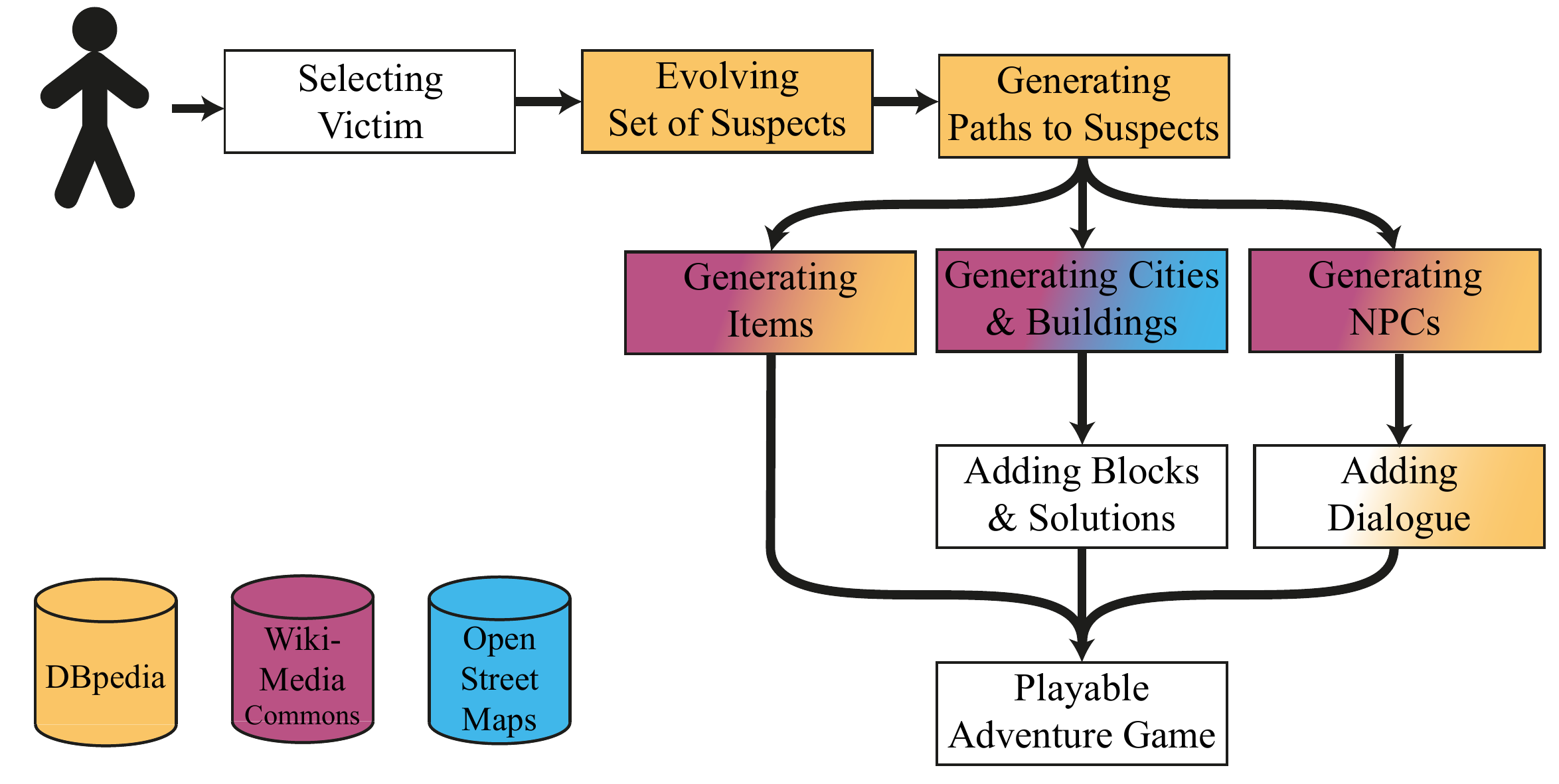}
\caption{Flowchart of WikiMystery and its open data sources.}
\label{fig:flowchart}
\end{figure}
The game generation involves several steps, as shown in Figure~\ref{fig:flowchart}, the first of which is selecting the victim. From the victim, the system uses DBpedia\footnote{DBpedia is a project which extracts information from Wikipedia in a structured manner (\url{http://dbpedia.org/}).} to find a set of five suspects via artificial evolution presented in Section \ref{sec:suspects} and to generate paths between each suspect and the victim via a constructive algorithm presented in Section \ref{sec:paths}. Once all paths are generated, the system creates locations, items and NPCs through constructive processes covered in Section \ref{sec:enriching}. Finally, it generates puzzles for accessing locations and dialog options for learning about clues or general information from NPCs; we discuss the latter in Section \ref{sec:dialogue}.


\section{Crawling DBpedia}

The plot of \emph{WikiMystery} is created from a series of hyperlinks in Wikipedia, generated using several consecutive queries to DBpedia. Once a victim has been introduced, the system tries to find
suspects and pinpoint a culprit among them. Then, it searches for paths between the victim and the suspects.


\subsection{Finding Suspects and Culprit}\label{sec:suspects}

\begin{figure}[t]
\centering
\includegraphics[trim={45px 600px 115px 0px},clip,width=\columnwidth]{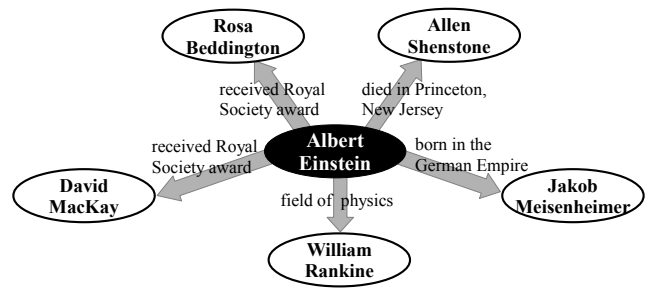}
\caption{Selecting suspects from DBpedia to use in the game and finding related between victim and suspects. Initially, the system has a single node: the victim (black node). Suspects related to the victim are selected with a genetic algorithm (white nodes), and paths between the victim and suspects are created from DBpedia (see Fig.~\ref{fig:paths}). All suspects share direct connections with the victim, shown on the arrows.}
\label{fig:radial}
\end{figure}

Selection of a set of suspects involves identifying who is related to the victim, and out of those, which subset is the most interesting. Given a Wikipedia article about a person, the system queries DBpedia to find anyone who has something in common with the victim. It can be as common as living in the same place, or as specific as being in the same band. This list is our pool of suspects. For each one, we query DBpedia to find everything known about them. At this point, we have a list of suspects, each containing a list of characteristics. Each characteristic can have multiple values. For example, a suspect could be ``Albert Einstein'', who would have the characteristic ``Field'' with values ``Physics'' and ``Philosophy''. Figure~\ref{fig:radial} shows a simplified selection of suspects. From a victim (black node), the system finds suspects (white nodes), who are related somehow to the victim (arrows). The system must also find a set of characteristics that can single out the culprit among suspects. A characteristic and a value of that characteristic, together, form an \emph{evidence of innocence}, which is used to identify the culprit and issue an arrest warrant.

The list of possible subjects, characteristics and values can be very large, at times. Characteristics may have multiple values available, of which we will only use one per characteristic, and suspects can have multiple characteristics. For example, as of the publication date of this paper, there are 15,300,451 distinct people related to Albert Einstein in some way, constituting his possible suspect pool. Each one has, on average, at least five characteristics, which may or may not have multiple values. Selecting five suspects, four characteristics and their values is therefore challenging. To select which subset of this list is interesting, we turn to an $\mu+\lambda$ evolutionary algorithm. Our goal is to have a finite set of suspects $n$ (typically, 5) and a finite set of characteristics $n-1$, such that we can pair each characteristic to a person. The leftover person is the culprit, and the characteristics are evidence of innocence for the $n-1$ suspects. This allows the player to eliminate innocent suspects by finding the clue paired to them. The remaining suspect who does not have the same value with any evidence of innocence must be the killer.

Our fitness function evaluates solvability and diversity. \emph{Solvability} favors complete solutions, where the player can identify the culprit by excluding the $n-1$ characteristics she knows the killer does not have.
Depth-First Search is applied: for every chromosome, it marks one of the suspects in the chromosome as the killer. The search states are characteristics in the chromosome, and they are paired with one suspect once they are visited. Valid states have three properties: (a) the killer has at least one value for the characteristic; (b) one or more suspects have value(s) for the characteristic; (c) at least one suspect has one value different from the killer. The algorithm tries to pair each suspect to one characteristic if the suspect has at least one value that is different from that of the killer (for that specific characteristic). If it cannot match a pair, it backtracks and tries a different suspect. No characteristic can be paired with more than one suspect, and vice-versa. The optimal solution is a leaf where all characteristics are paired successfully to suspects; the fitness is the depth of the leaf.

\begin{figure*}[htb]
\centering
\includegraphics[trim={25px 490px 25px 35px},clip,width=\textwidth]{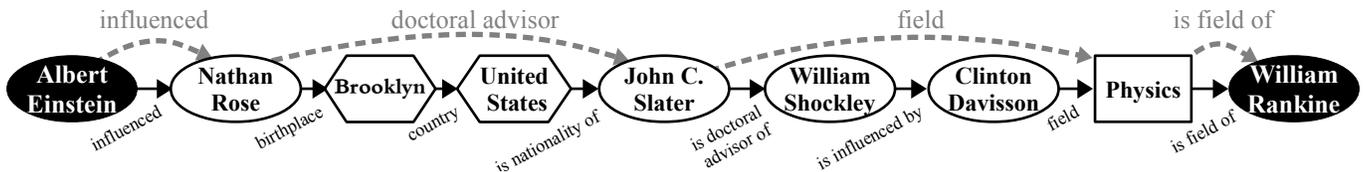}
\caption{The major and minor paths between Albert Einstein and William J. M. Rankine. Major paths have dotted arrows, minor paths have black arrows. Locations are represented as hexagons, NPCs as circles and items (books or photographs) as squares.}
\label{fig:paths}
\end{figure*}

\emph{Diversity} evaluates how different the characteristics and values are. The game only outputs one value per characteristic/suspect pair, so it is necessary to optimize which value to use. For example, in a game with 5 suspects and 4 characteristics, a solution with 20 values, one per pair, is better than one where one suspect has no value for any characteristic but the one matched to him. Additionally, a solution where all suspects have the same job and live in the same city is less diverse than a solution where they all have different jobs and live in different cities. Even though they still use the same characteristics (``job'' and ``residence''), the second one has more diversity of values. The actual fitness value is given by:
\begin{align}
f_{D} &= \sum_{i=0}^{P} \left ( Q_i \times \left(- \sum_{j=0}^{V_i}{ p_{ij} \times {log_2(p_{ij})} }\right)  \right) \label{eq:1} 
\end{align}
{\noindent}where $P$ is the number of characteristics; $V_i$ is the number of values for characteristic $i$; $Q_i$ is the number of people with characteristic $i$. We multiply with $Q_i$ to reward more suspects sharing characteristic $i$. $p_{ij}$ is calculated as the number of people that have value $j$ in characteristic $i$ divided by $Q_i$.

The system uses cascading elitism \cite{togelius2007towards} over a population of $100$ individuals for $500$ generations, with a mutation chance of 20\%. Cascading elitism uses both fitness functions: it sorts the population using the solvability fitness, removes the worse 50\% individuals, and then sorts the remaining using the diversity fitness. The highest 25\% of the population is duplicated and mutated until the new population is filled. It is far more important that games are solvable for playability's sake (rather than diverse); solvability is applied first during cascading elitism so that it introduces a stronger genetic bias.


\subsection{Finding Paths to Suspects}\label{sec:paths}

Once the system has a victim, suspects and clues as evidence of innocence,
it weaves them into a plot by searching DBpedia for a path of hyperlinks between the victim and each suspect. A path consists of nodes (Wikipedia articles) and edges (links between them). The set of all paths can be seen as a tree if we merge the initial node in the paths (i.e. the victim). Therefore the root of the tree is the crime scene, and each branch leads to a possible suspect. This tree represents most plot points the player is able to unravel in the game in order to move the plot forward, such as locations, NPCs and clues. The only clues not present in this tree are those of evidence of innocence.

For each suspect, the system queries DBpedia multiple times, searching all possible paths between the victim and said suspect. It rates each path based on how diverse it is, i.e. the type of articles and links in the path. For example, a path that only has articles about locations is less diverse than one with an even number of articles about people and locations. This process can be computationally expensive, since it is necessary to create one query per node in the path, and per direction of the edge. In practice, using paths longer than 5 nodes has proven to be too time consuming \cite{barros2016playing}. To bypass this, we divided the search into two steps. The first finds a path of length no longer than 5 nodes, as described above, which we call the \emph{major path}. Then, for each consecutive pair of nodes in the major path, the system searches for a \emph{minor path} between those nodes. The minor path replaces the edge between the two nodes in the major path. Figure \ref{fig:paths} shows an example path between the victim (Albert Einstein) and a suspect (William J. M. Rankine),  identifying major and minor paths.

The system measures path quality based on its \textit{length} and \textit{uniqueness}. Longer paths are preferred because they will extend the game: each node will be transformed into a city, NPC or item in the story. Uniqueness is calculated as the entropy of each node/edge in the path, compared to all nodes and edges found in all paths in that particular search. Thus a path where the type of edge is not found in other possible paths is better than typical edges (e.g.~for scientists it is typical to have edges of type ``influenced by'' or ``influenced'').


\section{Enriching the Data}\label{sec:enriching}

The system transforms the set of paths (tree) obtained from Wikipedia into gameplay objects that the player can interact with. Each node in the tree becomes a location, an item or a NPC. To do so, it creates all necessary game objects, then generates dialogs and links between them, verifies if all objects appear in the correct order, and add puzzles.

Nodes in the tree can be roughly categorized into places (e.g. ``London'' or ``Canada''), people (e.g. ``Albert Einstein'') and everything else (e.g. ``Mathematicians of the 20th Century''). The system begins by transforming the nodes into the simplest objects possible: locations, NPCs and items. For each node based on an article about a place, it generates a city (if the place contains a geographic coordinate) or a building. In the game's logic, the world contains cities, and buildings are places inside cities. Buildings can also contain items and characters. If the system generates a building, it tries to place it into its respective city. If it cannot find any city related to that building, it will randomly pick a place from Wikipedia and generate a city for it, placing the building in it.

After buildings and cities are created, the system takes all nodes based on real people and generates one NPC for each. The NPC gets the original person's name and a small description. Any node that is not a person or a location is transformed into an item: either a book, a list, a letter or a photograph. Depending on the type of item, different text templates are generated to explain it.

It is not possible to transform the tree's root into an NPC, because he/she is supposed to be murdered. WikiMystery attempts to solve this by adding people related to the victim instead. For each suspect, it searches for a person directly connected to the victim, and transforms him/her into an NPC. If it cannot find enough people, it generates ``random'' NPCs, whose sole purpose is to give a clue about the following node.

Once all objects needed for the plot have been generated, it is necessary to create a logical sequence of steps from the victim to each suspect. The system traverses each branch in the tree, and adds clues and conditions from one node to the next. If the current node is a location, an NPC or item is generated and placed in it. If it is a person, dialog is created directing the player to the next node. We discuss dialog generation in more detail in Section ~\ref{sec:dialogue}. Otherwise, the clue is added to the item's text description. Additionally, at random times the game may generate a ``fake'' NPC, the sole purpose of which is to provide a red herring. It is given a random name, no description, and dialog that is less than helpful. A condition manager guarantees that non-root game objects are only available after they have been triggered by another object.

Finally, the system adds ``puzzles''. One of the most well-known puzzles in adventure games is the ``lock-and-key'', i.e. a location that is inaccessible unless the player uses some specific item to unlock it. WikiMystery generates this kind of puzzles, creating items that are able to unlock buildings, such as flashlights for dark places and crowbars for chained gates. 
Puzzle objects are placed via a variation of the Breadth-First Search algorithm. First, nodes in the tree are separated by their depth, so depth of 0 will have only the location of the root NPCs, depth of 1 would contain all locations available after talking to the root NPCs, and so on. We simulate a playthrough to perform said separation. It also maintains an array with all possible keys (e.g. ``keys'',``crowbars'', etc) and an initially empty stack of locks. At every depth, it randomly chooses whether to put a key in a building of that depth. If it does so, it adds the respective lock to the stack of locks. Additionally, it may randomly pop a lock from the stack and add it to another building. For example, at depth 0 it may chose to put the ``flashlight'' key in the root building. It will automatically add the lock ``darkness'' to the stack. Because the victim's house is the only building at depth 0, and it has already been chosen, the algorithm goes to depth 1.
It randomly chooses to put no key and no lock, so it skips straight to depth 2, where it finds a Church and a House. It randomly decides to not put any keys, but decides to pop the lock from the stack (i.e. the ``darkness'' lock) and adds it to the Church. Adding the key before the lock guarantees that the puzzle will be solvable.


\section{Dialog Generation}\label{sec:dialogue}

The game's dialog has two goals: to advance the game by giving hints and evidence needed to win, and to provide a sense of depth and immersion, which can be hard to capture in a data game. Each NPC has their own dialog tree, i.e. the lines of dialog that both the player and the NPC use when interacting, along with the dialog options for the player. The root of this tree is a simple ``Hello'', and the choices that follow are called dialog branches. There are two types of branches: the \emph{main branch} containing information necessary to complete the game, and the \emph{side branch} (of which there are several subtypes), which contains information that is not necessary to complete the game but increases immersion.

\subsection{Main Branch}
The main dialog branch contains hints that will allow the player to advance in the game. For every NPC, the generator parses data and stores information from this person's DBpedia page: anything from places, persons, items, and concepts associated with the person, as well as personal information like birthplace and birthday, is stored. The dialog generator then takes sentence templates and replaces placeholder text. For example, Rosa Beddington's person object might have ``Jamaica'' stored within it as an associated place. When a player talks to another NPC who is associated with Rosa Beddington, they might have a dialog node telling the player where they think Rosa Beddington is.
\begin{displayquote}\small
I saw \texttt{PERSON} in \texttt{PLACE}. You should probably look there.
\end{displayquote}
will now become:
\begin{displayquote}\small
I saw Rosa Beddington in Jamaica. You should probably look there.
\end{displayquote}

After packaging a sentence of dialog into a dialog node, the generator adds this node as a child of the dialog root. Dialog choices are hidden by default unless its parent has been visited. Thus, the player must select the root ``Hello'' option before any branches are revealed to them.

\subsection{Side Branches}

\begin{figure}
\includegraphics[width=\columnwidth]{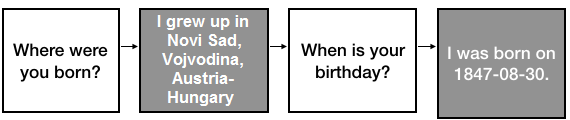}
\caption{An example of a dialog side branch. The player is speaking to Hermann Einstein in the Albert Einstein game.}
\label{fig:example_side_dialogue}
\end{figure}
Beyond the main dialog branch, the generator also selects randomly from a set of ``side branches'', which have no effect on the overall story. These branches provide extra information about the NPC that the player is speaking to. There are future plans to use these for an educational purpose, where players can learn more about characters' backgrounds by talking to them.
Currently, there are 3 possible side branches: birth-dates and birth-places, current residence, and lifetime achievement.
When data is originally parsed from DBpedia, birth, current residency and overview information are stored. When creating a side branch, this data replaces the placeholder words in templates (as in the main branch); see Fig.~\ref{fig:example_side_dialogue} for an example side branch.
After the main branch is created, the generator randomly selects up to two topics for which to generate side branches, or none. If generated, the system shows an option on the dialog screen that leads the user to these branches. 

\section{Example Playthrough}\label{sec:playthrough}

As an indicative playthrough, we describe the first few minutes of WikiMystery gameplay; this game uses as input the text ``Albert Einstein'', identified by the TIME magazine \cite{time1998einstein} as the ``Person of the [20th] Century''. Once the game launches, the user can load any of the 100 most influential people of the 20th century, which were pre-generated for the purposes of the analysis of Section \ref{sec:evaluation}.

\begin{figure}[t]
\centering
\subfloat[Initial World Map]{\includegraphics[trim=5px 5px 5px 25px,clip,width=0.48\columnwidth,height=0.09\textheight]{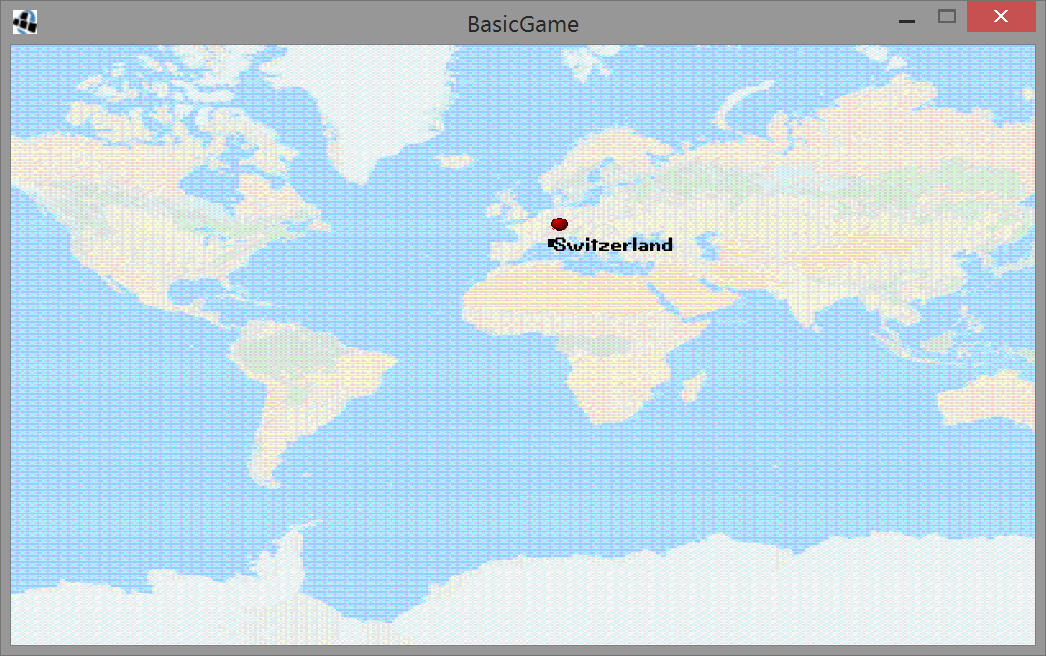}\label{fig:playthrough_2}}~
\subfloat[Initial city map of Switzerland]{\includegraphics[trim=5px 5px 5px 25px,clip,width=0.48\columnwidth,height=0.09\textheight]{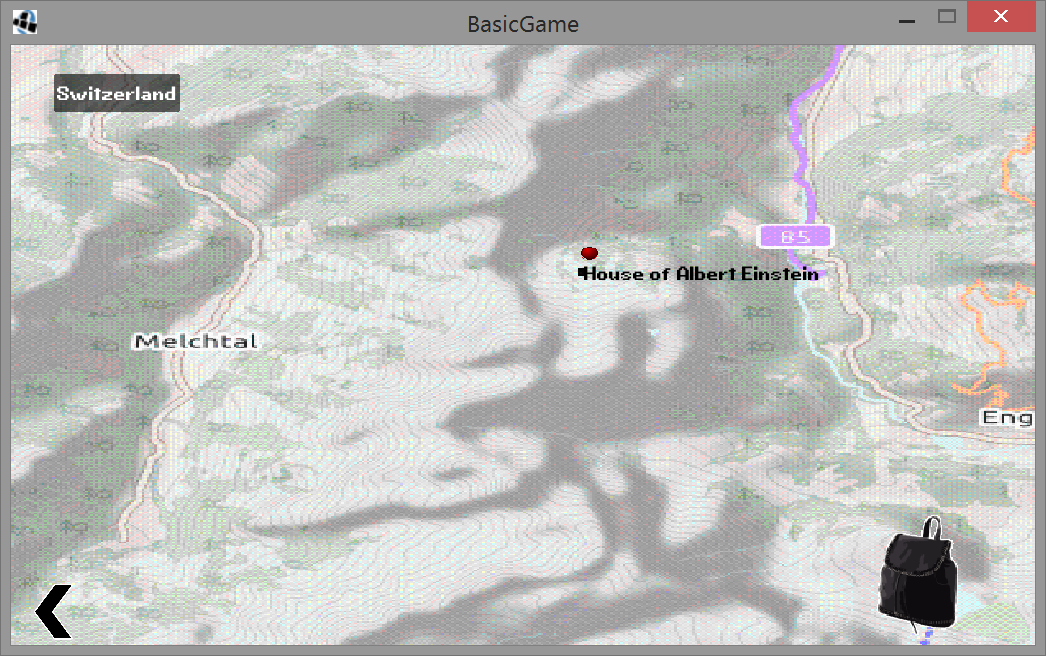}\label{fig:playthrough_3}}
\\
\subfloat[House of Albert Einstein]{\includegraphics[trim=5px 5px 5px 25px,clip,width=0.48\columnwidth,height=0.09\textheight]{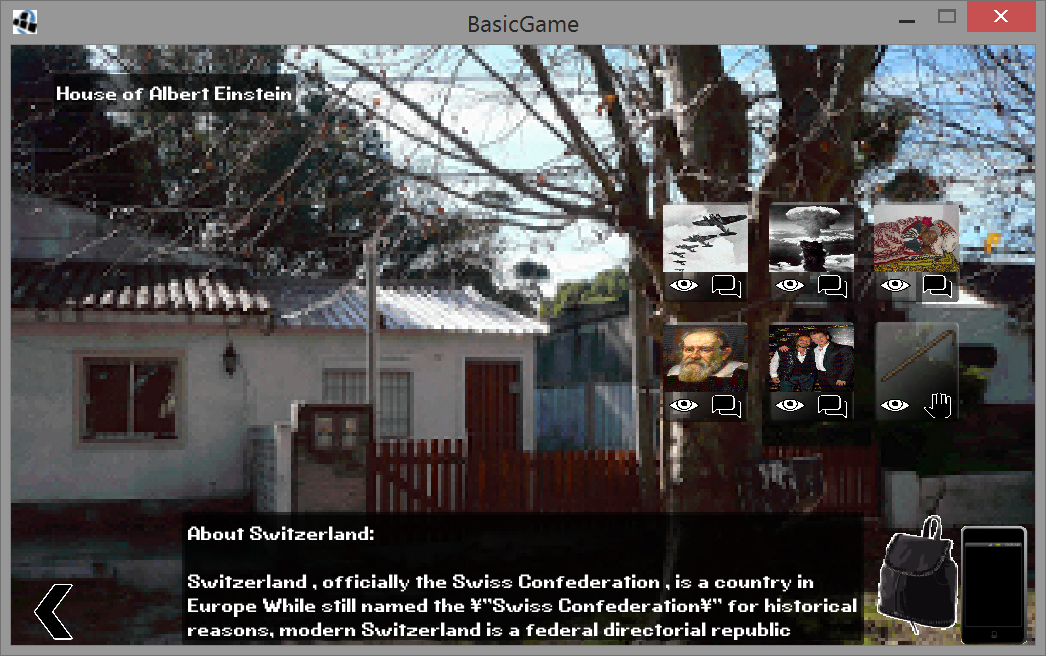}\label{fig:playthrough_4}}~
\subfloat[Clue given during dialog]{\includegraphics[trim=5px 5px 5px 25px,clip,width=0.48\columnwidth,height=0.09\textheight]{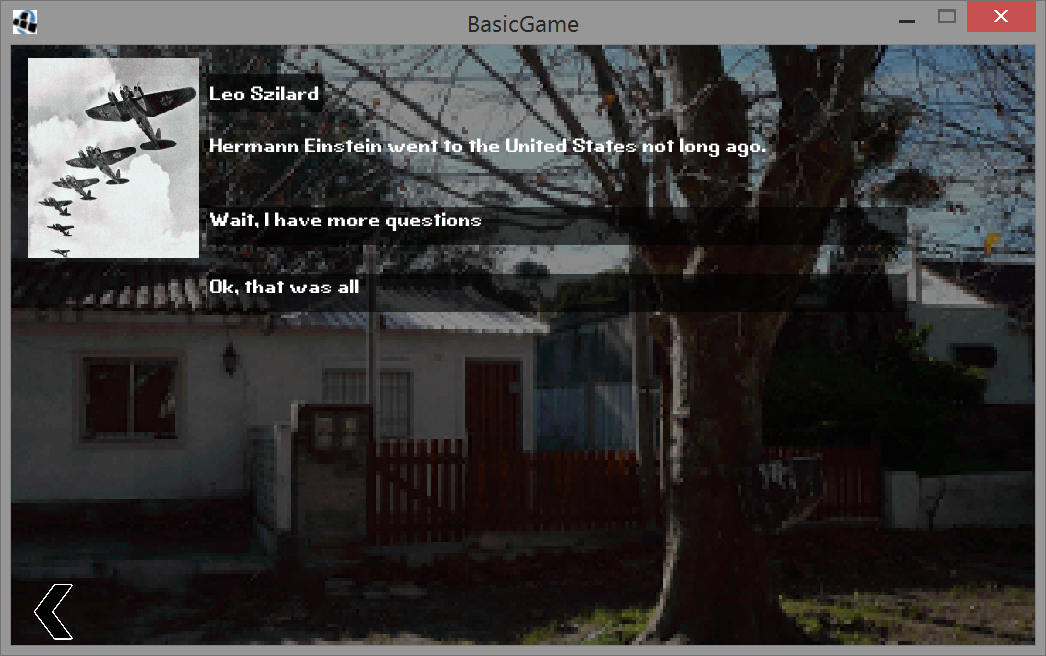}\label{fig:playthrough_5}}
\\
\subfloat[Travel between cities]{\includegraphics[trim=5px 5px 5px 25px,clip,width=0.48\columnwidth,height=0.09\textheight]{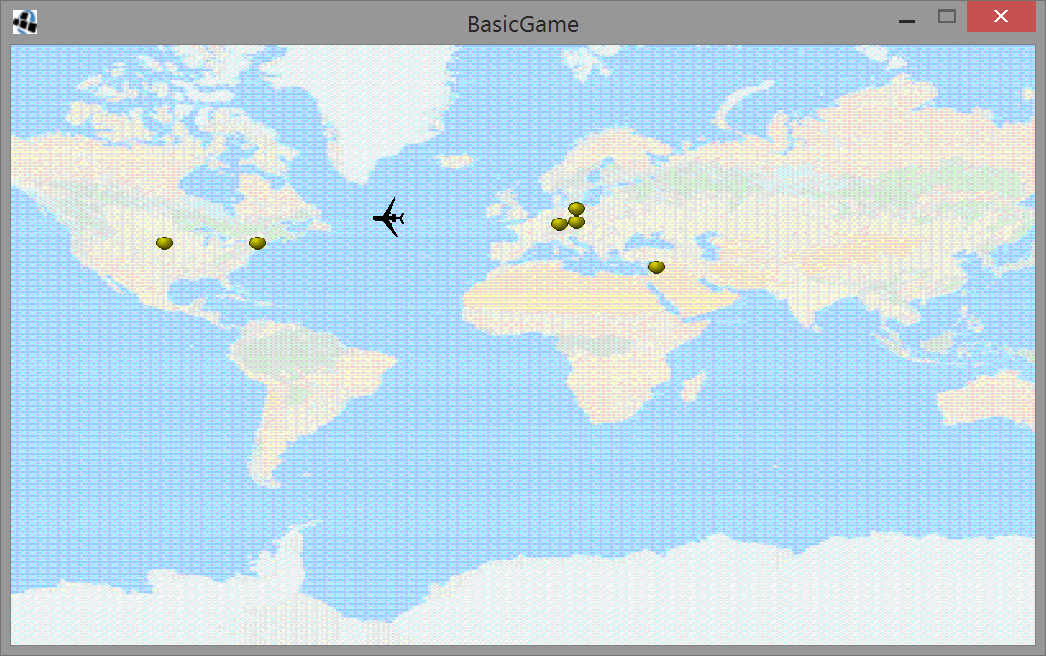}\label{fig:playthrough_6}}~
\subfloat[Icon of the Photograph item]{\includegraphics[trim=5px 5px 5px 25px,clip,width=0.48\columnwidth,height=0.09\textheight]{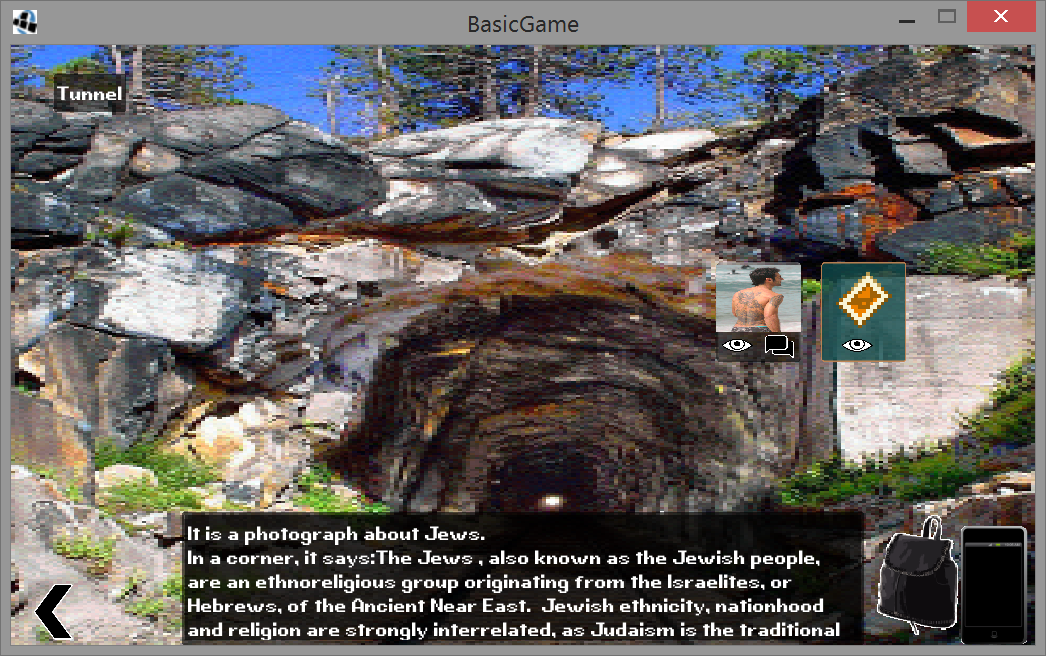}\label{fig:playthrough_7}}
\\
\subfloat[Zoomed-in Photograph item]{\includegraphics[trim=5px 5px 5px 25px,clip,width=0.48\columnwidth,height=0.09\textheight]{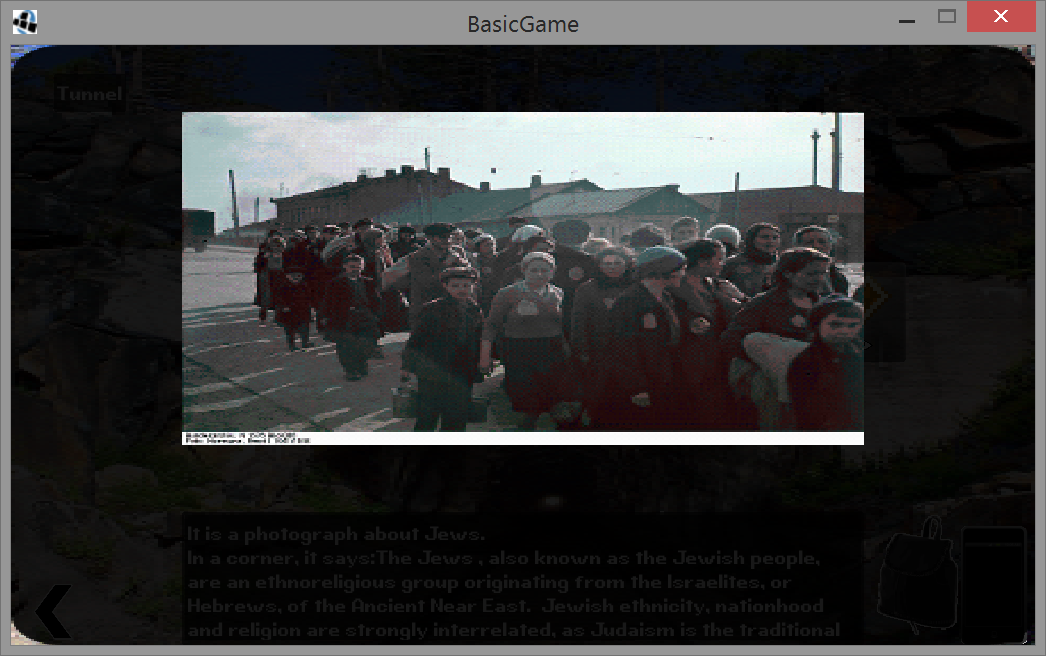}\label{fig:playthrough_8}}~
\subfloat[Dialog with a Suspect]{\includegraphics[trim=5px 5px 5px 25px,clip,width=0.48\columnwidth,height=0.09\textheight]{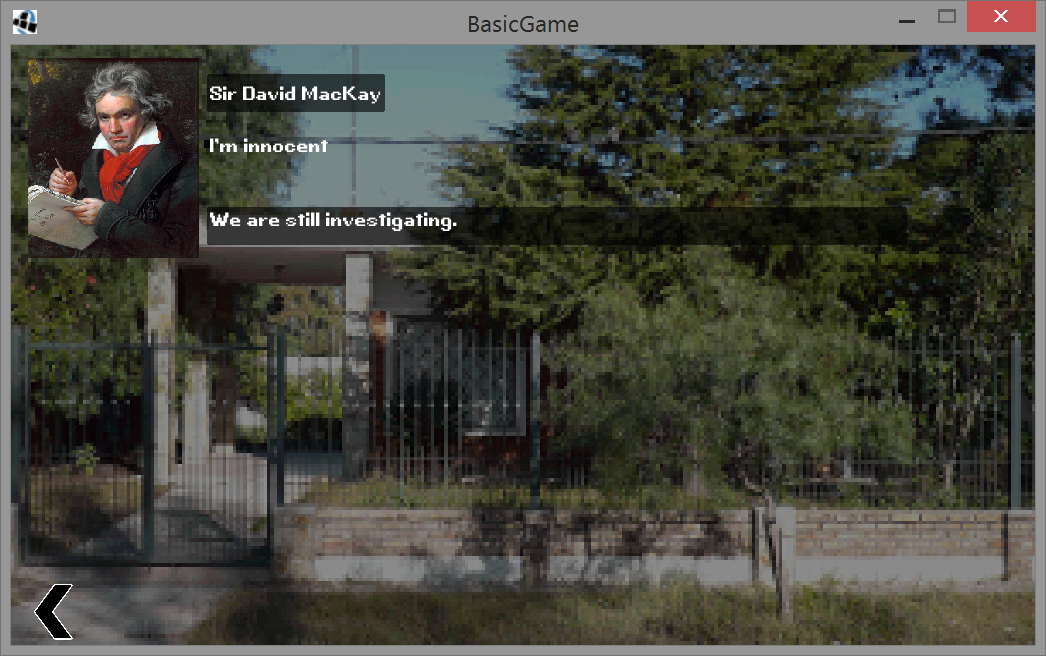}\label{fig:playthrough_9}}
\\
\subfloat[Finding an Evidence of Innocence]{\includegraphics[trim=5px 5px 5px 25px,clip,width=0.48\columnwidth,height=0.09\textheight]{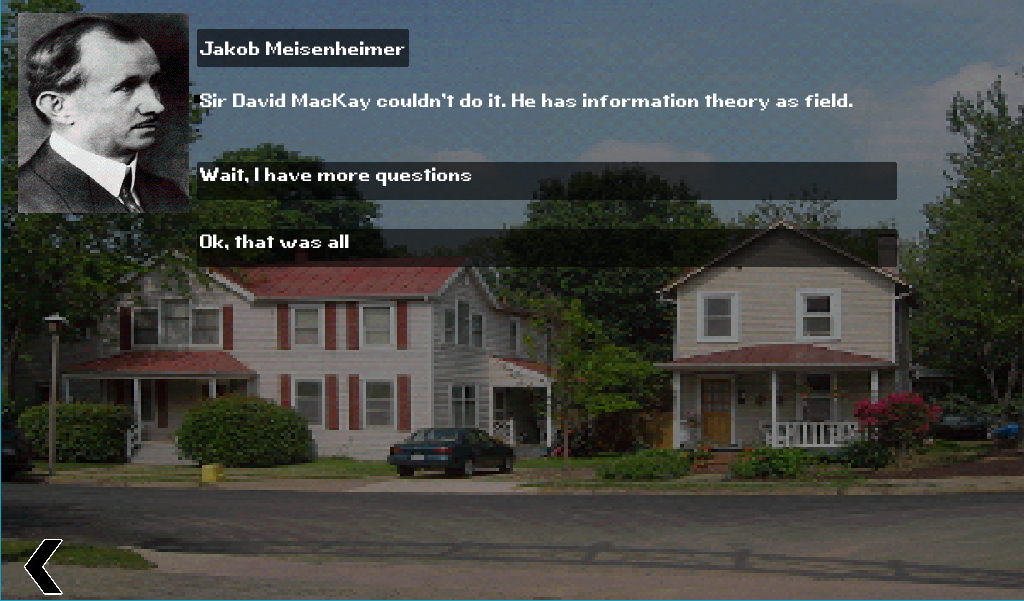}\label{fig:playthrough_11}}~
\subfloat[Issuing a warrant]{\includegraphics[trim=5px 5px 5px 25px,clip,width=0.48\columnwidth,height=0.09\textheight]{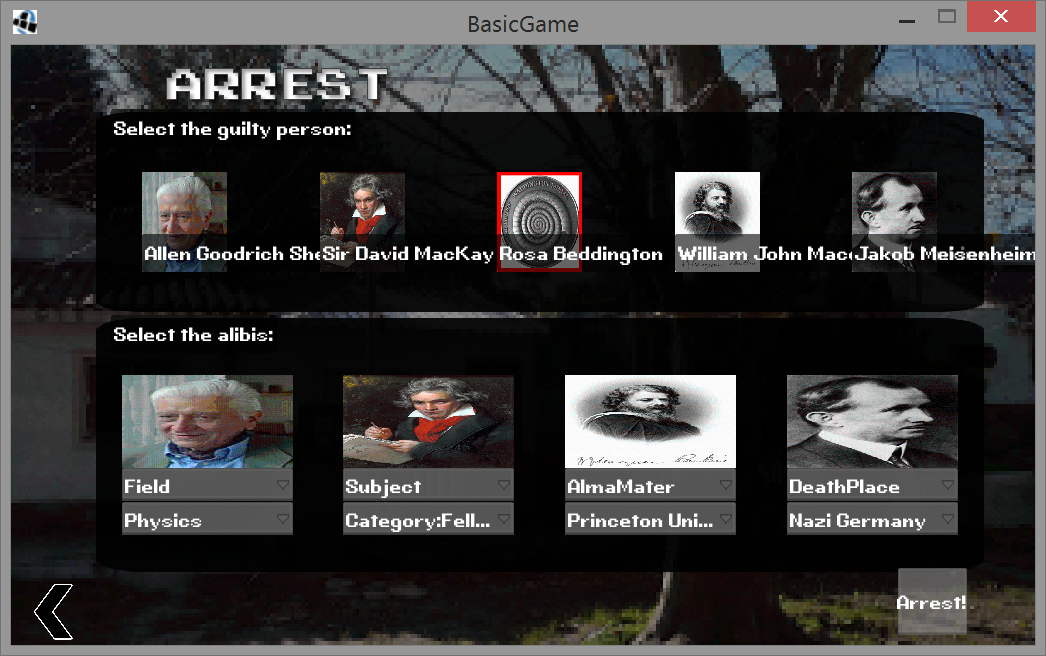}\label{fig:playthrough_10}}
\caption{In-game screenshots of the mystery around the murder of Albert Einstein.}
\label{fig:playthrough}
\end{figure}

The game starts at the world map (see Fig.~\ref{fig:playthrough_2}), where only one point can be visited: Switzerland, chosen as the birthplace of Albert Einstein. Clicking on that point of interest, the user moves to a map of a location in Switzerland collected from OpenStreetMap\footnote{Apparently what is shown in Fig.~\ref{fig:playthrough_3} is an open area near the Melchtal Valley, as the DBpedia entry places the coordinates of the country of Switzerland at its center.}, where a single location titled ``House of Albert Einstein'' (see Fig.~\ref{fig:playthrough_3}) can be visited. The player also has access to a backpack in this screen (bottom right of Fig.~\ref{fig:playthrough_3}), which is currently empty but can store items that can be used to access locked locations. When the player clicks on the house of Albert Einstein, they move to the building screen which shows a background of a single-story house, coupled with informative text about Switzerland in the bottom area (see Fig.~\ref{fig:playthrough_4}) and six different game icons that can be interacted with to the right. The first five icons are NPCs, while the last icon displays a crowbar which can be stored in the inventory by clicking on the hand button on the crowbar icon.

As noted above, in the House of Albert Einstein there are five NPCs which the player has the option of observing (eye button under each icon in Fig.~\ref{fig:playthrough_4}) or talking to (dialog button under each icon in Fig.~\ref{fig:playthrough_4}). These NPCs are Leo Szilard, David Joseph Bohm, Jean Gebser, Riazuddin, and James; the last NPC was randomly generated and given a random name, while the remaining NPCs are physicists except Gebser who is a philosopher. Clicking on the eye button gives information about each of these NPCs (for the random NPC James the text says ``There is no information available for this character''). Note that the images chosen for all NPCs are not images of people in the case of Szilard and Bohm (instead the images are related to the atomic bomb), and not images of the correct people in the case of Gebser and Riazuddin.

Clicking on the dialog button of any NPC icon moves the user to the dialog overlay (see Fig.~\ref{fig:playthrough_5}), where the NPC's name is followed by their response text, followed in turn by a set of dialog options. The general dialog sequence for any NPC in the house of Albert Einstein revolves around first asking for help, then asking their name, and then about any information they might have and where the player should look for it. The responses of each NPC depend on which path they are on, and towards which suspect they will guide the player. Indicatively, if the player initiates dialog with Leo Szilard then the NPC will respond to the player's question ``Please state your name.'' with ``Maybe... You can call me Leo Szilard. I was one influence.''; in this case, Leo Szilard was an influence to Albert Einstein, which (subtly) explains why this character is in this game. When the player asks ``Is there something you think I should know?'', Leo Szilard responds ``He talked to Hermann Einstein a lot.''; the player can then ask ``Where is Hermann Einstein?'' to which Leo Szilard responds ``Hermann Einstein went to the United States not long ago.'' (see Fig.~\ref{fig:playthrough_5}). This immediately adds ``the United States'' as a location in the world map, puts the house of Hermann Einstein on its city map, and an NPC named Hermann Einstein within it.

After the player has talked to all five NPCs in the House of Albert Einstein, there are five more locations in the world map that they can visit by clicking on them, at which point a small plane will be shown traveling from the player's current location to the selected one (see Fig.~\ref{fig:playthrough_6}). These locations on the world map are `the United States'\footnote{The marker based on DBpedia information is again placed at the center of the U.S.A.} (containing the house of Hermann Einstein), `Princeton, New Jersey' (containing a Tunnel building), `Israel' (containing the house of Nathan Rosen), `Wrttemberg' (containing a Stadium building) and `Swiss Federal Institute of Technology in Zurich' (containing a building of the same name, and placed in Zurich on the map). Similarly to the house of Albert Einstein, there is one or more NPCs or clues in each building listed above. For instance, in the Tunnel building of Princeton, New Jersey there is a random NPC named Vlad and a photograph icon (left-most in Fig.~\ref{fig:playthrough_7}). Clicking on the photograph shows an image of Jewish people (see Fig.~\ref{fig:playthrough_8}), and its description says: 

\begin{displayquote}\small
It is a photograph about Jews.

In a corner, it says: The Jews, also known as the Jewish people, are an ethnoreligious group originating from the Israelites, or Hebrews, of the Ancient Near East. Jewish ethnicity, nationhood and religion are strongly interrelated, as Judaism is the traditional faith of the Jewish nation, while its observance varies from strict observance to complete nonobservance.

There are names written behind:

\noindent Canada

\noindent Israel
\end{displayquote}

This information is based on the abstract of the Wikipedia article\footnote{\url{https://en.wikipedia.org/wiki/Jews}, accessed 13 March 2017.} regarding the `category: Jews' (as stored in DBpedia) which is used to link different NPCs in the mystery together. These NPCs are Nathan Rosen in Israel and a random NPC (also named Vlad) in the University building in a world map location named `Canada'; Vlad reveals that Allen Goodrich Shenstone is located in his house in Princeton, New Jersey.

After an extensive investigation taking the player to many different cities around the globe, and slowly revealing more and more buildings, NPCs and clues in previously visited cities, the player finds the five suspects. In this mystery the suspects are Sir David MacKay (whose dialog and pleas for innocence are shown in Fig.~\ref{fig:playthrough_9}), Allen Shenstone, William John Macquorn Rankine, Jakob Meisenheimer, Rosa Beddington. Each of these five names are also provided --- after questioning --- by one of the five NPCs in the player's starting location: the house of Albert Einstein. Of those suspect NPCs, Rosa Beddington (linked to Einstein as a fellow scientist and having been awarded by the Royal Society) is the culprit. Other suspects such as Sir David MacKay can be absolved by finding an evidence of innocence, in this case provided by chemist Jacob Meisenheimer (see Fig.~\ref{fig:playthrough_11}). Once the player is confident they have collected enough evidence, they can click on the cellphone (bottom left corner of Fig.~\ref{fig:playthrough_5}) to choose the guilty person as in Fig.~\ref{fig:playthrough_10}. The player chooses the guilty person and once they do so, the remaining suspects are placed in another window (bottom half of Fig.~\ref{fig:playthrough_10}); the player must then specify one characteristic and the correct value for each person which make them incapable of having committed the murder. The characteristics and values that absolve all suspects except Rosa Beddington are included in Table \ref{table:killers}. If the player selects the culprit and chooses values for the remaining suspects, they can click on the ``arrest'' button (bottom right of Fig.~\ref{fig:playthrough_10}) at which point the game ends with a message of success or failure.

\section{Evaluation}\label{sec:evaluation}

While the playthrough of Section \ref{sec:playthrough} provides a glimpse of what it means to play a generated murder mystery, this section evaluates the content generated from a broader set of murdered Wikipedia persons. The goal is two-fold: estimate the number of interactions afforded in each game (e.g.~dialogs with NPCs, visits to cities, item pickups), and assess the sensitivity of the system to different inputs (i.e. Wikipedia persons). For the former, several metrics regarding instances of specific elements (cities, NPCs, dialog lines) per generated game are listed. For the latter, we describe which Wikipedia persons were murdered in games with the highest and lowest values in these metrics. While this paper does not perform an user playtest of such generated games to assess e.g.~how intuitive the connections between NPCs are, the provided evaluation is vital in understanding how complex the generated games are and which of the generated gameplay elements contribute most to this complexity. This evaluation is thus a first step prior to a playtest, to assess for instance the minimum number of player clicks (via the tree size metric combined with the dialoge nodes metric) for a game to be completed. Such metrics can then be compared with actual metrics derived during playtests, but can also inform changes to the generative algorithms before such playtests can take place.

To assess a broad range of games, based on persons with a strong presence in Wikipedia, we used the list of the TIME magazine's 100 most influential people of the 20th century \cite{time1998einstein} as input. Each person in the list became the victim in a procedurally generated game, some after preprocessing, excluding two: ``American G.I.'' and ``Unknown Rebel''. The system was not able to generate games with them, as the first represents a whole category (we could not choose a single person that represented this category), and the latter represents an unknown person who does not contain the tag ``Person'' in his DBpedia page. Additionally, the system cannot process groups of people, so inputs such as ``The Kennedy Family'' had to be transformed into a single individual. Entries about groups were transformed into one of the most known people of the group. For example, ``The Beatles'' became ``John Lennon'' and ``The Kennedy Political Family'' became ``John F. Kennedy''.

The system generated a total of 98 games, one per input. Table ~\ref{table:averages} shows the quantitative results.

\begin{table}
\centering
\caption{Average metrics of all generated adventure games for the 98 most influential people.}
\label{table:averages}
\begin{tabular}{|l|c|}
\hline
\multicolumn{2}{|c|}{Location metrics}\\
\hline
Cities						&18.07 \\
Buildings					&46.37 \\
Average buildings per city 	&2.88 \\
\hline
\multicolumn{2}{|c|}{Item \& Puzzle metrics}\\
\hline
All Items					&23.91 \\
Books						&9.45\\
Photographs	(torn or not)	&7.24\\
Torn Photographs			&2.55\\
Key items					&2.71\\
Locked buildings 			&2.67 \\
\hline
\multicolumn{2}{|c|}{NPC metrics}\\
\hline
All NPCs					&46.53 \\
NPCs based on real people	&24.03 \\
Average ratio of real NPCs over all NPCs &52\%\\
Average NPCs per building 	&1.02\\
\hline
\multicolumn{2}{|c|}{Dialog metrics}\\
\hline
All dialog nodes				&208.33\\
Average dialog nodes per NPC	& 4.45 \\
All side-branches 				&35.47 \\
Achievement side-branches & 9.05\\
Residence side-branches & 8.56\\
Birth side-branches & 17.86\\
\hline
\multicolumn{2}{|c|}{Complexity}\\
\hline
Average length of paths & 12.07 \\
Tree size & 60.34\\
\hline
\end{tabular}
\end{table}

\subsection{Game Content}

Based on Table \ref{table:averages}, the average tree size of the generated games is 60.3 nodes. The game with the smallest tree size had ``Robert H. Goddard'' as input and 20 nodes. ``Marlon Brando'', ``Martin Luther King, Jr.'', ``Richard Rodgers'' and ``Willis Carrier'' tied for the most nodes in the tree, with 65. On average, the length of paths between victim and each suspect was 12 nodes.  
Six games had the lowest path length with 5 nodes, while 39 had the highest path length with 13 nodes.

Each game had on average around 18 cities and 46 buildings, with approximately 2.9 buildings per city. The most common cities amongst all games was ``The United States'', appearing in 73 out of the 98 games, followed by ``New York City'' (43) and ``District of Columbia'' (37). North American locations dominated the top 10 most common cities, with 8 locations. The remaining two were ``London'' and ``Germany''. Note that while the game only represents locations as cities and buildings, the in-game city category may include countries (e.g. The United States), states and actual cities.

On average, 24 items were generated per game, mostly books (9.45). Games with most and fewest books were created from, respectively, ``Le Corbusier'' (20 books) and ``Theodore Roosevelt'' (1). Key items and locked buildings tend to appear together, with an average of 2.71 key items and 2.67 buildings. Every game had at least one key and one locked building, while at most there were three keys and three locked buildings in a single game. The number of keys was always equal or higher to that of locked buildings, ensuring solvability.

An average of 46 NPCs were created per game, and on average 24 were based on real people (ratio of 52\%). While this ratio is not optimal, we believe it can be improved in future versions by being more lenient in the NPC generation: now, we only look at people with one-degree distance from the article that originated the node. If there is no person, we could expand the search to 2 or 3 degrees distances, which we believe can improve this ratio. We believe that increasing the percentage of NPCs based on real people over ``random'' NPCs would provide more interesting, full-fledged characters and interactions. The ratio of NPCs based on real people ranged from  28\% to 76\% of all NPCs. The game with most real NPCs was generated from ``Lech Wa\l{}esa'' with 41 real NPCs, and the one with the least from ``Walter Reuther'' with 9.
 
Based on Table \ref{table:averages}, there are 208.33 dialog nodes on average in a WikiMystery game, distributed across all NPCs in the game. Results show an average of 4.45 dialog nodes per person. Every person has a main branch in their dialog tree, so the number of main branches is equal to the number of NPCs. In addition to those, there are on average 35.5 side-branches in a game. 
Of those, an average of 9 side-branches refer to the NPC's personal achievements, 8.6 side-branches concern the NPC's current residence, and 17.9 side-branches are associated with the person's birth. There are nearly twice as many side-branches on birth than the other two types, since the generator creates two branches (birth date and birth place) when selecting a side-branch on birth.   

\begin{table*}
\centering
\caption{Solution from games generated with the top 3 most influential people: Albert Einstein, Franklin D. Roosevelt and Mahatma Gandhi. Each innocent suspect is paired to one characteristic (in blue italics) that will differentiate him/her from the killer (whose name is shown with an asterisk, and is last in each list). Empty values appear in the game as ``Unknown''. The direct connection column shows the primary criteria for choosing the suspect in relationship to the victim (e.g. ``politicians'' means that both the victim and that suspect were politicians. thus the suspect was directly connected to the victim).}
\label{table:killers}
\begin{tabular}{ l |  
>{\centering\arraybackslash} m{2cm}
>{\centering\arraybackslash} m{2.5cm}
>{\centering\arraybackslash} m{2.5cm}
>{\centering\arraybackslash} m{3cm}
|
>{\centering\arraybackslash} m{3cm}
}

\multicolumn{6}{c}{\textbf{\large{Albert Einstein}}}\\
\hline\hline
\textbf{
Suspects} & \textbf{Death Place} & \textbf{Field} & \textbf{Subject} & \textbf{AlmaMater} & \textbf{Direct Connection}\\
\hline
Jakob Meisenheimer & \textcolor{blue}{\emph{Nazi Germany}} &   & 1934 deaths & Ludwig Maximilian University of Munich & Born in the German Empire \\
\hline
Sir David MacKay &   & \textcolor{blue}{\emph{Information theory}} & Living people & California Institute of Technology & Received the Royal Society award\\
\hline
William J. M. Rankine & Glasgow & Physics & \textcolor{blue}{\emph{Thermodynamicists}} & University of Edinburgh & Physicists \\ 
\hline
Allen G. Shenstone & United States & Physics & Fellows of the Royal Society & \textcolor{blue}{\emph{Princeton University}}  & Died in Princeton, New Jersey \\
\hline
Rosa Beddington* & Great Tew & Developmental biology & 20th-century women scientists & Brasenose College, Oxford & Received the Royal Society award \\
\multicolumn{5}{c}{}
\\[0.2em]

\multicolumn{6}{c}{\textbf{\large{Franklin D. Roosevelt}}}\\
\hline \hline
\textbf{
Suspects} & \textbf{Term end} & \textbf{Party} & \textbf{Office} & \textbf{AlmaMater} & \textbf{Direct Connection}\\
\hline
Kevin Cahill & \textcolor{blue}{\emph{1994-12-31}} & Democratic Party (US) from the 103rd District & Member of the New York Assembly at New Paltz & State University of New York & Part of Democratic Party\\
\hline
Gwendolyn Garcia & 2013-06-30 & \textcolor{blue}{\emph{One Cebu}} & Governor of Cebu & University of the Philippines Diliman & Politicians\\
\hline
Johnny Ellis & 1993-01-11 & Democratic Party (US) & \textcolor{blue}{\emph{Majority Leader of the Alaska Senate}} & Claremont McKenna College  &  Part of Democratic Party, politicians\\
\hline
Jane Griffiths & 2005-04-11 & Labour Party (UK) & Member of parliament & \textcolor{blue}{\emph{Durham University}} & Politicians \\
\hline
Daniel Poulter* & 2015-05-12 &  & Member of parliament & University of Bristol & Politicians \\
\multicolumn{5}{c}{}
\\[0.2em]

\multicolumn{6}{c}{\textbf{\large{Mahatma Gandhi}}}\\
\hline \hline
\textbf{
Suspects} & \textbf{Death year} & \textbf{Birth place} & \textbf{Subject} & \textbf{Occupation} & \textbf{Direct Connection} \\
\hline
Tex Avery & \textcolor{blue}{\emph{1980}} & Taylor, Texas & Articles containing video clips & Animator, cartoonist, voice actor, director & Appear in the Wikipedia category of articles containing video clips\\
\hline
Stanley Rosen & 2014 & \textcolor{blue}{\emph{Cleveland}} & Jewish American writers &  & 20th century philosophers\\
\hline
Volker Zotz &  & Landau & \textcolor{blue}{\emph{20th-century philosophers}} & Writer & 20th century philosophers\\
\hline
Jhunnilal Verma & 1980 & Damoh, India  & People from Damoh & \textcolor{blue}{\emph{Lawyer}} & Indian lawyers\\
\hline

Eddie Lyons* & 1926 & Beardstown, Illinois, USA  & Articles containing video clips & Actor, director, screenwriter, producer & Appear in the Wikipedia category of articles containing video clips\\

\end{tabular}
\end{table*}

\subsection{Suspects, Direct Connections and Evidence}

Each generated game must have a set of suspects, evidence of innocence and direct connections between suspect and victim (i.e. the reason for selecting those suspects). Table ~\ref{table:killers} shows the set of suspects, evidence of innocence and direct connection between victim and suspect of the three most influential people on TIME's list: Albert Einstein, Franklin D. Roosevelt and Mahatma Gandhi. Values in italics are used as evidence for innocent suspects, allowing the player to differentiate between them and the culprit. Notice that when the culprit has no value (e.g. in the game generated from Franklin Roosevelt, Daniel Poulter has no value for the ``Party'' characteristic), any value would fit to differentiate between any suspect, but the game will only check this characteristic for the specific paired suspect (in the game generated from Franklin Roosevelt, Gwendolyn Garcia is paired up with the ``Party'' characteristic).
Additionally, if two innocent suspects share the same value for a characteristic, it is used as evidence for one of them, as long as it is different from the culprit. That merely means that it is evidence that only one suspect is innocent, but does not absolve the other suspect. An example is shown in Mahatma Gandhi's game (see Table ~\ref{table:killers}) where Tex Avery and Jhunnilal Verma both died in 1980, but this is only evidence of innocence for Avery (note that Eddie Lyons, the culprit, died in 1926).
In some cases, the actual value appeared as a consequence of Wikipedia's own organization. In Mahatma Gandhi's game, the primary reason for selecting Tex Avery and Eddie Lyons as suspects is them belonging (like Gandhi) to the category ``Articles containing video clips'', indicating that they both appear in the Wikipedia list of articles containing video clips. The secondary reason was that this set of these five suspect allowed for a solvable and somewhat diverse game, according to the GA.

Direct connections are relations between the victim and each suspect. Since they depend on hyperlinks from the victim's Wikipedia page, they can only be as varied as the article itself. Therefore, games usually have an emergent underlying theme. In the game generated for Albert Einstein, both he, Allen Shenston and William J. M. Rankine were in the field of Physics, and he and Shenston both also died in Princeton, New Jersey; Einstein, David Mackay and Rosa Beddington received the Royal Society award, and both Einstein and Jakob Meisenheimer were born in the German Empire.
In Franklin D. Roosevelt's game, Roosevelt and all suspects except Kevin Cahill are politicians\footnote{Cahill actually \textit{is} a politician, he is not tagged as one in DBpedia.}. Roosevelt, Cahill and Johnny Ellis were all part of the U.S.A. Democratic Party. Finally, in Mahatma Gandhi's game, both he and Jhunnilal Verma were Indian lawyers; Gandhi, Volker Zotz and Stanley Rosen were 20th century philosophers; and he, Eddie Lyons and Tex Avery all appear in the Wikipedia category of articles containing video clips. The last connection is arguably poorer than others, demonstrating how some of the source data can be difficult to tailor to our needs; this is discussed extensively in Section \ref{sec:discussion}.

\section{Discussion}\label{sec:discussion}

The sample playthrough of Section \ref{sec:playthrough} and the numerical evaluations of Section \ref{sec:evaluation} provide a high-level overview of the types of games generated by the current {WikiMystery} prototype. Contrary to our early attempts at adventure generation, which created one path between two people \cite{barros2015dataadventures}, the murder mystery is far less linear and includes more dialog and gameplay options. The fact that paths can be traversed non-sequentially (inevitably so, as it is difficult to keep track of which NPC or object forms a path towards which suspect) increases the exploration and branching factor in terms of decision-making on the part of the player. This in turn leads to more interesting gameplay as it gives a greater sense of player agency. The gameplay has been improved with more branching, better visual presentation of results, more interesting dialog options, and a concrete winning condition.

While it was a priority for the authors to improve on the gameplay quality of the broader Data Adventures project, the biggest appeal remains its link to real-world data accessed via open data repositories. Based on the metrics of Table \ref{table:averages}, that aspect of WikiMystery has been strengthened as well, with each game containing a multitude of cities placed in their real-world locations and with a city map showing their street view (based on OpenStreetMap). The ratio of random NPCs to ``real'' NPCs, which are based on Wikipedia articles, is also kept in balance, while the introduction of photograph in-game objects increases the modes through which open data can be experienced (i.e. through images rather than text information found in book in-game objects). Most importantly, the improved NPC dialog allows not only for a more engaging and intuitive way to solve the mystery but also allows for yet another way to present open data, as a player can choose what questions to ask of the NPC (e.g. regarding their life achievements) rather than being presented the data as a large chunk of text when observing the NPC, for example. 

\begin{figure}
\centering
\subfloat[Darling house of early colonial Australia highlighted... and placed in Israel.]{\includegraphics[trim=5px 5px 5px 25px,clip,width=0.48\columnwidth,height=0.09\textheight]{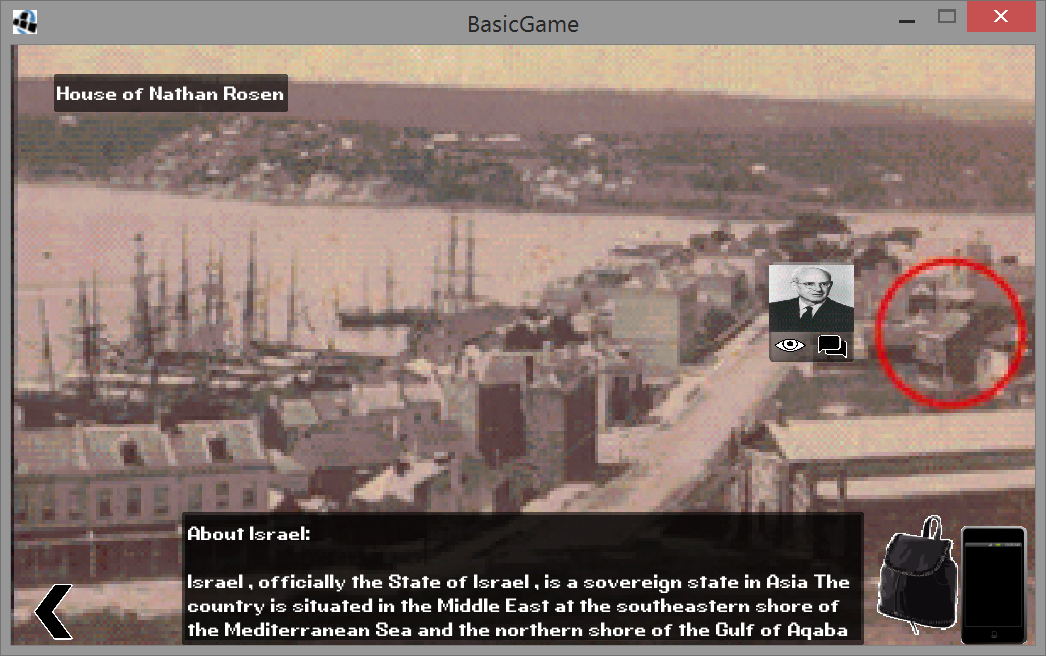}\label{fig:bug_1}}~
\subfloat[The image of Confucius chosen for Hermann Einstein, and the  `ask about Jews' dialog are... unfortunate.]{\includegraphics[trim=5px 5px 5px 25px,clip,width=0.48\columnwidth,height=0.09\textheight]{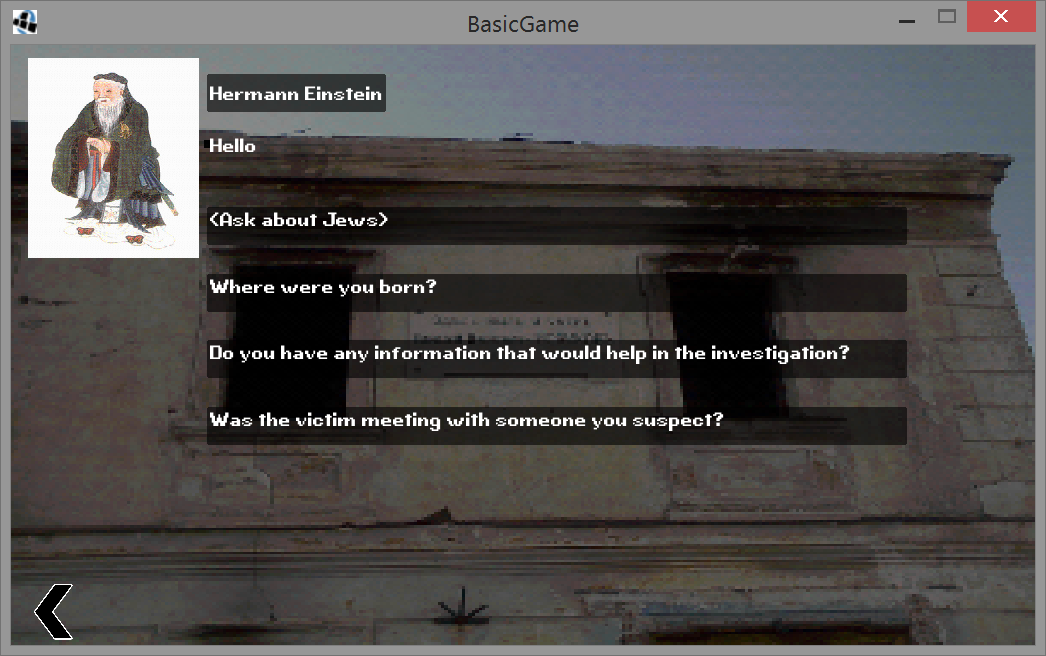}\label{fig:bug_2}}
\caption{Absurd and potentially offensive combinations of data can occur with WikiMystery.}
\label{fig:playthrough}
\end{figure}

Although there have been substantial improvements in the presentation of content from earlier iterations of Data Adventures \cite{barros2015dataadventures,barros2016playing}, the very nature of generating games from open data hinges on the uncontrollable nature of such data. This allows for nigh-infinite expressivity, as any person with a Wikipedia presence can potentially star in a generated game (where he is murdered), but this lack of control can lead to unexpected, unintended or even unwanted outcomes. On the one hand, the ongoing efforts of the Data Adventures line of research focus on controlling this vast repository of data and transforming it into intuitive and playable objects; for instance, attempting to find unique connections between people rather than trivial ones such as ``they are both human''. It is, however, impossible to ever fully control or constrain the experience, as doing so would obfuscate its origins from a living, vast knowledge base rooted deeply in the real world. It is that very absurdity that makes the outcomes appealing in their own way; as the user of another data-based game titled A Rogue Dream states, it feels ``like playing a videogame against The Internet'' \cite{cook2014aroguedream}, and at least in the case of WikiMystery that is intentional. 

This absurdity, however, causes some hilarious, and sometimes appalling, outcomes. It has been noted in the playthrough of Section \ref{sec:playthrough} that most NPCs' images were not correct, which is either due to the lack of appropriate images for those people in WikiMedia Commons, or flaws in the image parsers currently at hand. In such cases, a random search for an image of a man (for male NPCs) and a woman (for female NPCs) is used instead. For buildings, moreover, the image search is based on the name of the building without context of its geographical location. This can lead to results such as that of Fig.~\ref{fig:bug_1}, where not only is the building's background an old photograph with an actual highlighted building with a red circle, but on closer inspection the chosen building (result of a search for ``house of Nathan Rose'') is the Darling House, which holds historical significance for early colonial Australia, but in the game is used as a domicile for Nathan Rosen in Israel. Our choice of using only freely available sources, such as Wikimedia Commons, complicates the retrieval of specific images. A source such as Google Images could improve results, but contradicts the scope of freely available solutions. Therefore, future work should improve the search for appropriate images, possibly by increasing the breadth of searches in more repositories or by performing some computer vision verification that e.g. the image is one of a person.

Additionally, more problematic are instances where an unforeseen combination of content and their transformation can lead to insensitive or offensive results. As an example, Fig.~\ref{fig:bug_2} shows the dialog with Hermann Einstein as part of the playthrough of Section \ref{sec:playthrough} where the player is seeking the culprit of Albert Einstein's murder. The image, unsurprisingly, is not that of Hermann Einstein; instead, the random search for an image of a man serendipitously ended up being a drawing of Confucius. On the other hand, the dialog has chosen to highlight that the connection between this person and the next along the path is the `category: Jews'; this category was also cued by the photograph of Fig.~\ref{fig:playthrough_8} as discussed extensively in Section \ref{sec:playthrough}. In this case, the player interacts with Hermann Einstein with the dialog line ``$\langle$Ask about Jews$\rangle$''. It is certainly true that the actual story of Albert Einstein was deeply affected by him being Jewish and the events of World War II, so the category and the path found is accurate (perhaps desirable), however the random choice of this dialog line\footnote{Consider how inoffensive a similar line saying ``$\langle$Ask about Physics$\rangle$'' would be.} and the random assignment of an image of Confucius for avatar are a very unfortunate, insensitive and likely offensive combination. It is difficult to envision how such instances could be avoided, as it was largely an issue with simple transformations of data and their combination going awry. While it is not the case here, one should also not underestimate that the nature of open online data ``is often tainted by popular belief, misconception, stereotype and prejudice, as opposed to purely factual information'' \cite{cook2014aroguedream}, and thus such unfortunate instances may actually occur due to prejudice in the source data before they are even transformed. 

There are still important directions for future work, in order to improve both the usability of the game and its narrative consistency. For example, some interface additions, such as a travel diary, could help the player to keep track of clues and connections between NPCs, objects, locations and characteristics. Moreover, the current dialog format uses fixed templates and sequences; perhaps a grammar-based approach such as \emph{Tracery}~\cite{compton2015tracery} could result in more diverse, life-like dialogs. Furthermore, an important missing component in the narrative of the murder is the culprit's motive (and possible motives for other suspects). It is unlikely that a motive such as jealousy would be based on real facts and data, although it could be generated based on relationships of people (e.g. siblings or spouses) as in \cite{stockdale2016cluegen}. Exploring the relations between the NPCs, their personalities and goals seems promising. Using data to do so is not trivial, as even a sentiment analysis over a Wikipedia article about a person would only express the writer's feelings, not the actual subject's. On the other hand, we have yet to fully understand how players interact and view the data presented in WikiMystery. It is one of our priorities to release a playable version of the game online, setting up a logging system so that we can perform user studies. We also intend to investigate the possibility of using WikiMystery to gain insight on the correctness of data in DBpedia and Wikipedia.

\section{Conclusion}\label{sec:conclusion}

This paper presented the latest installment of the \emph{WikiMystery} game, and detailed its complex generation pipeline, from the name of a person with a Wikipedia article to a full interactive murder mystery game. Open data is used in a multitude of ways in order to find NPC suspects for an in-game murder of a specified person, to find paths linking these NPCs, to place them in locations around the globe and to provide a way for the player to absolve innocents and deduce the culprit. Moreover, open data is used to create the `levels' (i.e. cities and buildings) in which NPCs are found, to create in-game objects with photographs and books that act as clues, and to enhance dialog options of NPCs beyond the merely functional needs of completing the game. While there are many directions of future work in order to increase gameplay intuitiveness, to provide a better link between visuals and other content, and to reduce the absurdity of combinations, the current WikiMystery generator is the first to create fully playable adventure games with minimal human authorship and curation.

\section{Acknowledgments}
NPCs discussed in the generated adventures are instantiated from real people, but the similarities end there. The NPCs' actions in the game (as victims or culprits) in no way reflect the real-world people they are based on. The generator's output in no way accuses or misrepresents these real-world individuals. WikiMystery creates fictional counterparts of public figures who have a presence in Wikipedia:  any similarity between the (fictional) NPCs in the game and real-world people is therefore due to the data available in these open, online, freely accessible and editable repositories.

We  thank  Ahmed  Khalifa  and  Scott  Lee  for helpful insight. Gabriella Barros acknowledges financial support from CAPES and Science Without Borders program, BEX 1372713-3.

\bibliographystyle{IEEEtran}
\bibliography{IEEEabrv,wikiMystery}

\end{document}